\documentclass{article}
\usepackage[legalpaper, margin=1in]{geometry}
\usepackage{authblk, amsmath,epsfig, amsfonts, amssymb, xcolor, hyperref, caption, subcaption, graphicx, xcolor,comment}
\usepackage[linesnumbered,ruled,vlined]{algorithm2e}
\usepackage[utf8]{inputenc}
\usepackage[english]{babel}

\usepackage{algorithmic}
\newcommand{\argmin}{\operatornamewithlimits{argmin}}% Example definitions.
\DeclareMathOperator*{\argmax}{argmax} % thin space, limits underneath in displays% --------------------

\def\R{\mathbb{R}}
\def\L{{\cal L}}

\def\hx{{\hat x}}
\def\L{{\cal L}}
\def\R{{\mathbb R}}
\def\E{{\mathbb E}}
\newcommand{\Dt}{\mathcal{D}_{t}}

\date{}

\title{\mbox{Deep Diffusion Processes for Active Learning of Hyperspectral Images}}
\author[1]{Abiy Tasissa}
\author[2]{Duc Nguyen}
\author[1]{James Murphy\thanks{This research is partially supported by the US National Science Foundation grants NSF-DMS 1912737, NSF-DMS 1924513, and NSF-CCF 1934553.}}
\affil[1]{Department of Mathematics, Tufts University, Medford, MA}
\affil[2]{Department of Mathematics, University of Maryland, College Park, USA}

\begin{document}

\maketitle

\begin{abstract} A method for active learning of hyperspectral images (HSI) is proposed, which combines deep learning with diffusion processes on graphs.  A deep variational autoencoder extracts smoothed, denoised features from a high-dimensional HSI, which are then used to make labeling queries based on graph diffusion processes.  The proposed method combines the robust representations of deep learning with the mathematical tractability of diffusion geometry, and leads to strong performance on real HSI.  

\end{abstract}

%\textbf{Keywords}: hyperspectral images, variational autoencoders, deep clustering, active learning, semisupervised learning, diffusion geometry

\section{Introduction}
\label{sec:Introduction}
 
 Machine learning has provided revolutionary new tools for remote sensing, but state-of-the-art methods often require huge labeled training sets.  In particular, supervised deep learning methods can achieve near-perfect labeling accuracy on high-dimensional hyperspectral images (HSI), provided large libraries of labeled pixels are available \cite{Zhu2017_Deep}.  This hinders the practicality of these methods, as in many settings, data is collected at a pace that far exceeds human ability to generate corresponding labeled training data.
 
 In order to account for this, methods that require only a very small number of labels are needed.  The \emph{active learning} regime is particularly attractive for HSI labeling problems.  In active learning, an algorithm is provided with an unlabeled dataset, and the algorithm iteratively queries points for labels.  By choosing query points intelligently, the active learning algorithm can yield the classification performance of a much larger training set chosen uniformly at random.  
 
We propose an active learning method for HSI based on deep feature extraction and random walks on graphs.  First, an unsupervised variational autoencoder is used to nonlinearly denoise and compress the high-dimensional HSI.  Then, the resulting features are considered as vertices of a graph, and a Markov diffusion process on the graph is used to determine label queries and label all data points.  The proposed method combines the efficient feature learning of deep autoencoders with the mathematical interpretability of graph diffusion processes, and leads to strong empirical performance on real HSI.  

%%%
%%%

\section{Background}
\label{sec:vaes}

\subsection{Variational Autoencoders}
\label{subsec:vaes}

In an autoencoder architecture, input data is cascaded through nonlinear layers to obtain a latent representation. The latent representation is then cascaded  through nonlinear layers to obtain output data. These two stages respectively define the encoder and decoder. Typically, a loss function that enforces the reconstructed output to be similar to the input is minimized and the trained autoencoder learns
a low-dimensional latent feature useful for downstream tasks. In contrast to the autoencoder, in the variational autoencoder (VAE) \cite{kingma2014auto}, the output of the encoder is not a deterministic map but parameters of a distribution.  In particular, an encoding network maps $x \in \R^N$ and obtains parameters of the latent variable distribution $q(z|x)$. A latent feature $z$  sampled from this distribution is an input to a decoder that outputs $\hat{x}\sim p(x|z)$. A typical prior for the distribution of the latent variable is a Gaussian random variable $p(z)\sim N(0,I)$. Given this, the VAE optimization consists of two terms: (i) a reconstruction loss $\L_1=  \E_{z \sim q(z|x)} \log(p(x|z))$ that enforces that the reconstructed output $\hat{x}$ is similar to the input $x$; and (ii) a Kullback-Leibler divergence loss $\L_2 =  KL(q(z|x),N(0,I))$  that enforces that $q(z|x)$ agrees with $p(z)$. The encoder and decoder are jointly trained by maximizing the total loss $\L_1+\L_2$. 

\subsection{Learning by Active Nonlinear Diffusion}

The active learning algorithm employed in this paper is based on the ideas in \cite{Murphy2019_Unsupervised, Maggioni2019_LUND, Murphy2020_Spectral}.  In \cite{Maggioni2019_LAND}, the authors propose a semisupervised algorithm, learning by active nonlinear diffusion (LAND), that obtains the most important data points to query  for labels.  Important features of LAND are (i) it is a principled algorithm with provable performance guarantees; (ii) it accounts for nonlinear clusters possibly in high dimensions; and (iii) it is robust to noise and outliers \cite{Maggioni2019_LAND}.

We represent an HSI as $X=\{x_{i}\}_{i=1}^{n}\subset\mathbb{R}^{N}$ where each pixel is a point in $\R^{N}$ where $N$ is the number of spectral bands. Let $NN_{k}(x_{i})$ denote the set of $k$-nearest neighbors of $x_{i}$ in $X$ using the Euclidean distance metric.
The $n\times n$ weight matrix $W$ is defined as $W_{ij}=\exp(-\|x_{i}-x_{j}\|_{2}^{2}/\sigma^{2}), x_{j}\in NN_{k}(x_{i})$ with $\sigma$ denoting a scale parameter. With this, the notion of the degree of $x_{i}$ naturally follows as $\deg(x_{i}):=\sum_{x_{j}\in X}W_{ij}$. To define a random walk on $X$, we employ the $n\times n$ transition matrix $P_{ij}={W_{ij}}\big/{\deg(x_{i})}.$ It can be easily verified that $P$ has a spectral decomposition $\{(\lambda_{\ell},\Psi_{\ell})\}_{\ell=1}^{n}$. The \emph{diffusion distance at time $t$} between $x_{i},x_{j}\in X$ is defined as $D_{t}(x_{i},x_{j})=\sqrt{\sum\nolimits_{\ell=1}^{n}\lambda_{\ell}^{2t}(\Psi_{\ell}(x_{i})-\Psi_{\ell}(x_{j}))^{2}}$.  We note that $t$ tells us how long the diffusion process runs. In this paper, we use $t=30$ for experiments.

The main part of the LAND algorithm is to identify points to query for labels. LAND uses a kernel density estimator (KDE) and diffusion geometry for this task. In particular, the KDE is defined as $p(x)=\sum_{y\in NN_{k}(x)}\exp(-\|x-y\|_{2}^{2}/\sigma_{0}^{2})$ with $\sigma_0$ denoting a scale parameter.  For $x\in X$, let

\begin{align}\label{eqn:rho}
\rho_{t}(x) &=
\begin{cases}
\displaystyle\min_{p(y)\ge p(x), x\neq y} D_{t}(x,y), &x\neq \displaystyle\argmax_{z}p(z), \\
\displaystyle\max_{y\in X} D_{t}(x,y), & x=\displaystyle\argmax_{z}p(z),
\end{cases}
\end{align} 
be the diffusion distance to the nearest neighbor of higher density.  The maximizers of $\Dt(x)=p(x)\rho_{t}(x)$ are queried for labels.  These labels are propagated to other data points by proceeding from high to low density and assigning each unlabeled point the same label as its $D_{t}$-nearest neighbor of higher density that is labeled; see Algorithm \ref{alg:VALAND} and \cite{Maggioni2019_LAND} for details. 

\subsection{Related Work}
In recent years, deep generative methods, such as \emph{generative adversarial network (GANs)} and \emph{variational autoencoder networks (VAEs)}, have been used for feature extraction in many machine learning tasks \cite{ehsan2017infinite,makhzani2015adversarial}. In the context of clustering, a set of methods, known as deep clustering, propose learning features of the data and clustering simultaneously, showing strong empirical results \cite{tian2014learning,song2013auto,xie2016unsupervised}. For HSI images, several works have employed different deep learning architectures to extract essential features for downstream tasks such as classification \cite{chen2014deep,chen2016deep,li2017spectral,he2017multi,paoletti2019deep}. 

Active learning is a learning paradigm where the user has the ability to select the training data \cite{cohn1995active,mackay1992information}.  The underlying idea is that a few informative training samples could be sufficient for training an algorithm and obtaining accurate results. This framework has been used in remote sensing for HSI image classification \cite{liu2016active,wang2017novel,murphy2018iterative,tuia2009active}.  The main idea in this paper is that the active learning process depends on the representation and geometry of the data. The closest work to ours is  \cite{pourkamali2019effectiveness} where the authors combine active learning with VAEs. Therein, $K$-means clustering is first used to partition the space and then labels are acquired using uniform random sampling in each partition. Given the labels, a classifier is then trained in the latent space for the prediction task. One of the highlights of the proposed method is that the clustering algorithm LAND handles a broader class of cluster geometries than $K$-means does. 

We note that in contrast to the similar work \cite{li2020variational}, our method is in the active learning framework and the feature extraction and diffusion process via LAND are decoupled. 

\section{Proposed Algorithm}
\label{sec:ProposedAlgorithm}
We propose an active learning algorithm, VAE-LAND (see Algorithm \ref{alg:VALAND}), which has two main stages. The first stage is feature extraction of an unlabeled high-dimensional dataset using a VAE. The second stage employs the LAND algorithm to infer the true labels.  The proposed algorithm combines the power of VAEs to extract features with diffusion geometry on graphs to find impactful labels to query, which then propagate to other points.

\RestyleAlgo{boxruled}
\begin{algorithm}[!htb]
	\caption{\label{alg:VALAND}Variational Autoencoder Learning by Active Nonlinear Diffusion (VAE-LAND)}
	\flushleft
	\flushleft
	\textbf{Input:} $\{x_{i}\}_{i=1}^{n}$ (Unlabeled Data);  $t$ (Time Parameter); $B$ (Budget); $\mathcal{O}$ (Labeling Oracle)
	
	\vspace{3pt}
	
	\textbf{Output:} $Y$ (Labels)
	
		\vspace{3pt}

	\begin{algorithmic}[1]
	\STATE Run VAE on unlabeled data to obtain the latent representation $ \{\hx_{i}\}_{i=1}^{n}$.  
\STATE Compute $P$ and $\{(\lambda_{\ell},\psi_{\ell})\}_{\ell=1}^{M}$ using $ \{\hx_{i}\}_{i=1}^{n}$.
\STATE Compute kernel density estimate $\{p(\hx_{i})\}_{i=1}^{n}$ and $\{\rho_{t}(\hx_{i})\}_{i=1}^{n}$ (\ref{eqn:rho});
	\STATE Compute $\Dt(\hx_{i})=p(\hx_{i})\rho_{t}(\hx_{i})$.  
	\STATE Sort the data in decreasing $\Dt$ value to\ 
	acquire the ordering $\{\hx_{m_{i}}\}_{i=1}^{n}$.\FOR{$i=1:B$}
	\STATE Query $\mathcal{O}$ for the label $L(\hx_{m_{i}})$ of $\hx_{m_{i}}$.
	\STATE Set $Y(\hx_{m_{i}})=L(\hx_{m_{i}})$.
	\ENDFOR
	\STATE Sort $X$ according to $p$ in decreasing order as \\ $\{\hx_{\ell_{i}}\}_{i=1}^{n}$.  
	\FOR{$i=1:n$}
	\IF{$Y(\hx_{\ell_{i}})=0$}
	\STATE $Y(\hx_{\ell_{i}})=Y(z_{i}), \, z_{i}=\displaystyle\argmin_{z}\{D_{t}(z,\hx_{\ell_{i}})$\ $| \ $ $p(z)>p(\hx_{\ell_{i}}) \text{ and } Y(z)>0\}$.
	\ENDIF
	\ENDFOR
	\end{algorithmic}
\end{algorithm}

\section{Experimental Results}
\label{sec:Experiments}
\begin{figure}[ht]
\centering
\includegraphics[width=.23\textwidth,clip]{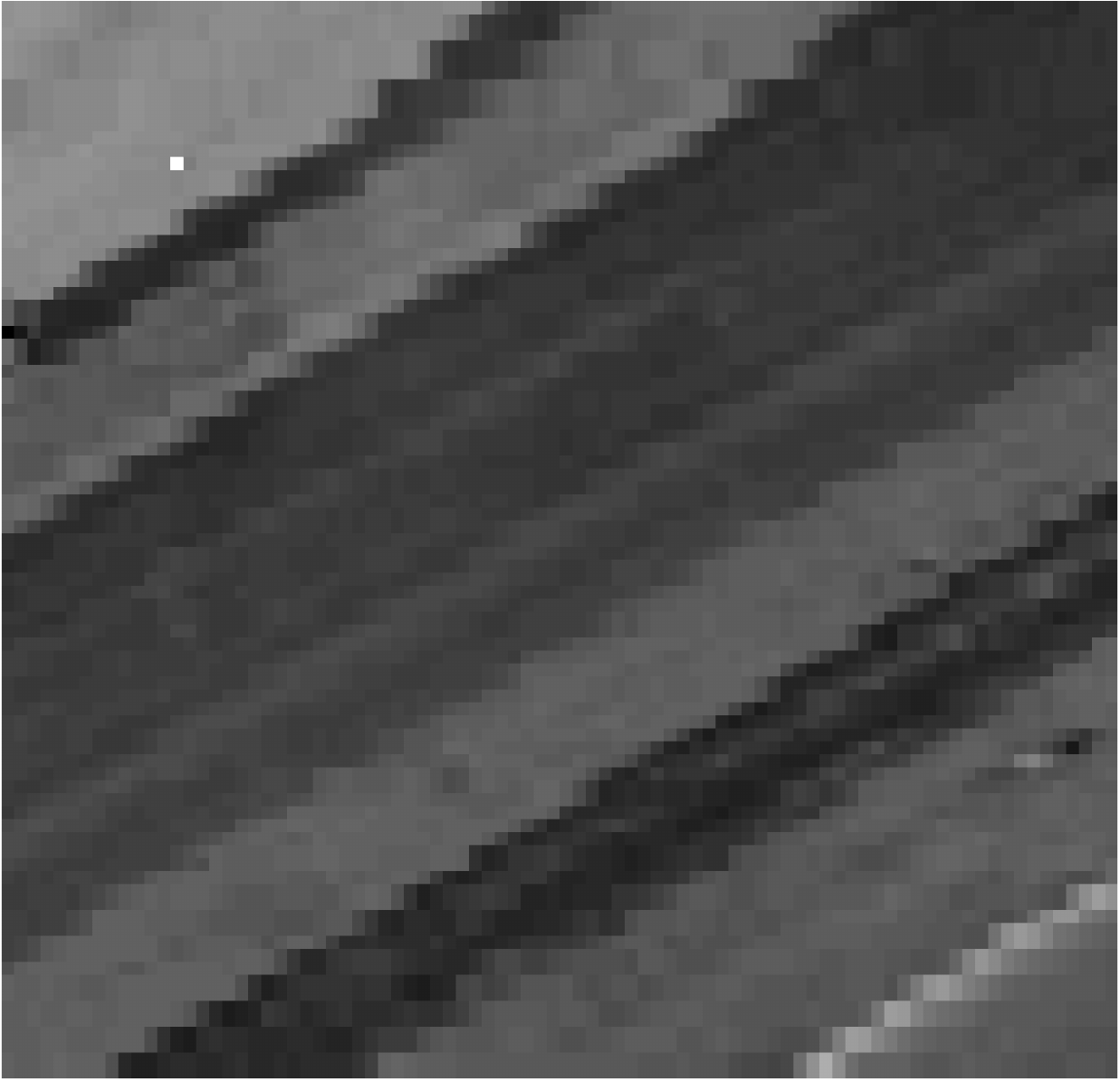}
\includegraphics[width=.23\textwidth,clip]{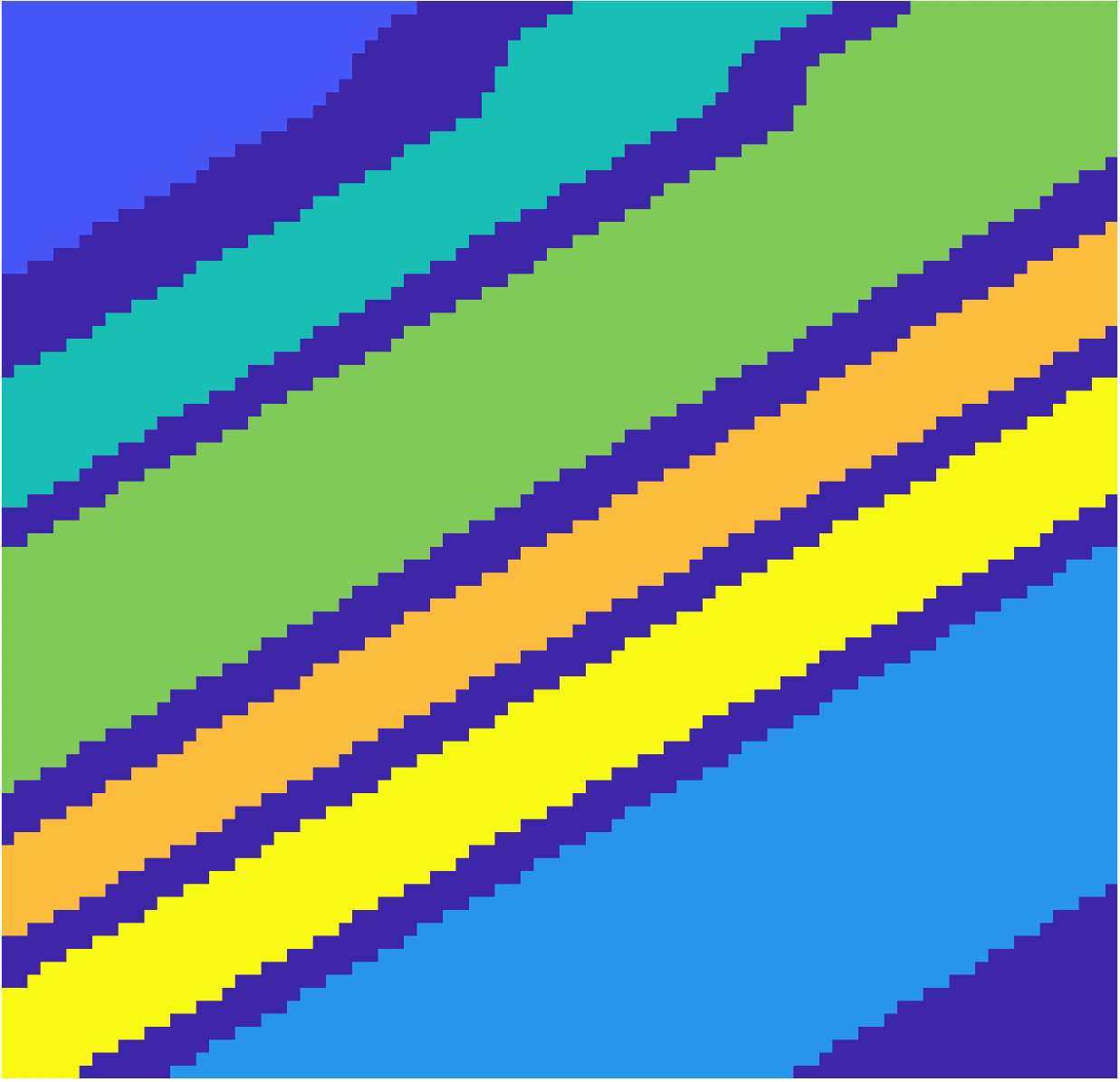}
\caption{\small{The $86\times 83$ Salinas A HSI data consists of 6 classes.  \emph{Left:} the sum of all spectral bands.  \emph{Right:} the ground truth.}}
\label{fig:SalinasA}
\end{figure}

We demonstrate the accuracy of the proposed algorithm experimentally.  The training of the VAE is done using Tensorflow in Python. For doing the active learning via LAND, we use the publicly available MATLAB code 
at  \url{https://jmurphy.math.tufts.edu/Code/}. Our code can be found at \url{https://github.com/abiy-tasissa/VAE-LAND}. Our test HSI dataset is the Salinas A hyperspectral dataset. The Salinas scene was captured over Salinas Valley, California. The image has a spatial resolution of 3.7-meter pixels and contains 224 spectral bands. The ground truth consists of 16 classes. We consider the Salinas A dataset, which is a subset of the Salinas dataset, and contains 6 classes. The Salinas A dataset and the ground truth data are publicly available (\url{http://www.ehu.eus/ccwintco/index.php/Hyperspectral_Remote_Sensing_Scenes#Salinas-A_scene}). Figure \ref{fig:SalinasA} shows a visual of the high-dimensional data and the ground truth labels. The performance of the algorithm is assessed using overall accuracy, defined as the ratio of correctly estimated labels to total number of labels after optimally aligning with the ground truth. 
\begin{figure}[ht]
\centering
\includegraphics[clip,width=.45\textwidth,trim=1cm 3.5cm 1cm 1cm]{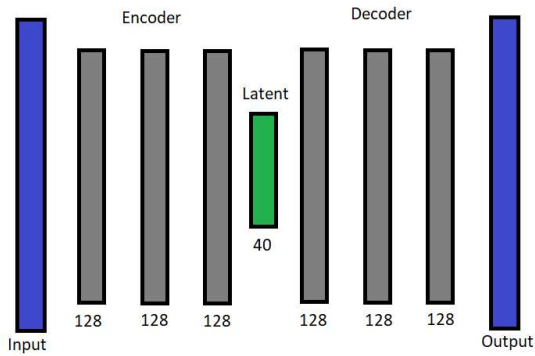}
\caption{\small{A schematic of the VAE architecture.  Input is a vector in $\R^{224}$. The encoder and decoder are fully connected neural networks. Both have three layers with 128 neurons in each layer. For all layers, the activation function is the rectified linear unit (ReLU).
The input is cascaded through an encoder. The extracted latent feature in $\R^{40}$ is then cascaded through the decoder to obtain the
output vector in $\R^{224}$.  Unlike the standard autoencoder, the extraction of the latent feature is not deterministic (see Section \ref{subsec:vaes} for discussion.)}}
\label{fig:vae_schematics}
\end{figure}
The Salinas A HSI dataset of size $83 \times 86\times 224$ is represented as a point cloud of size $7138\times 224$. We use the unlabeled data to learn a latent space representation of Salinas A in $\R^{40}$ dimensions. A schematic of the VAE architecture is shown in Figure \ref{fig:vae_schematics}. We optimize the VAE loss function using the Adam algorithm with learning rate set to $10^{-4}$. After training the VAE, we input the optimal latent space representation of the Salinas A dataset to the LAND algorithm for the task of inferring the ground truth labels of the HSI data. We compare our result to the standard LAND algorithm that labels the Salinas A dataset in its original representation. Since LAND is an active learning framework, we consider varying number of labeled data points ranging from $10$ to $2000$. In addition, we compare the active learning methods to query the samples with randomly selected training data. Figure \ref{fig:SalinasA} compares the performance of LAND and performance of VAE-LAND. First, for both VAE-LAND and standard LAND, LAND queries lead to significantly better accuracy than random queries. The proposed algorithm, VAE-LAND, attains an accuracy of 96.97\% with just 10 labeled points. This is a $12.5\%$ improvement to the accuracy of the standard LAND algorithm for the same number of labeled points. The standard LAND algorithm requires $400$ labeled points to reach accuracy of $90\%$ while for the same number of labeled points, VAE-LAND has an accuracy of $98.35\%$. 

\noindent \textbf{Complexity and run time}: The complexity of LAND is $O(C_{NN}+nK_{NN}+n\log(n))$ where $C_{NN}$ is the cost of computing all $K_{NN}$ nearest neighbours \cite{Maggioni2019_LAND}. 
The computational cost of VAE is difficult to estimate as it depends on several factors (e.g architecture, activation function, choice of SGD algorithm). In our numerical experiments, the cost of VAE is the dominating cost. Since LAND runs on low-dimensional features extracted from VAE, it is efficient. 

\begin{figure}[h]
    \centering
    \includegraphics[width=.45\textwidth]{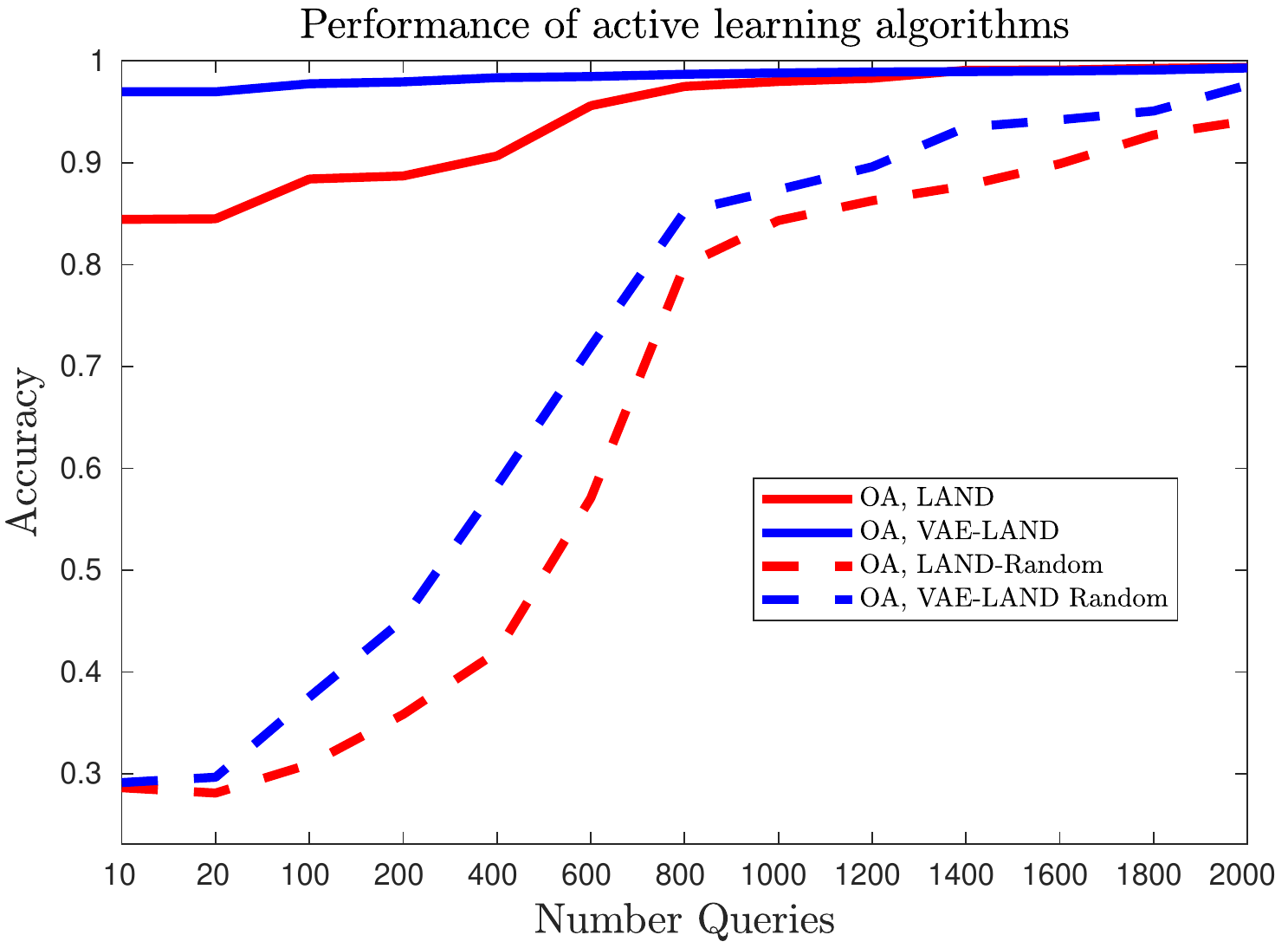}
    \caption{For the Salinas A dataset, the performance of VAE-LAND learning achieves a higher accuracy than the standard LAND algorithm. With just $10$ points, the overall accuracy of VAE-LAND is 96.97\%, a 12.5\% improvement to the competitive LAND algorithm. Both VAE-LAND and LAND obtain significantly better results than using randomly selected training instances. }
    \label{fig:my_label}
\end{figure}

\section{Conclusions and Future Directions}
\label{sec:Conclusions}
The proposed active learning algorithm, VAE-LAND, improves over the standard LAND and gives accurate results even when the number of queries are limited. The method uses VAE to generate good features, and uses the diffusion geometry-based LAND algorithm to determine query points. The LAND algorithm then uses these queried labels to predict the labels of the unlabeled data samples. In future work, we shall explore data models for which the algorithm has theoretical performance guarantees.

\bibliographystyle{IEEEbib}
\bibliography{DeepDiffusionHSI}

\end{document}